\documentclass[runningheads]{llncs}
\usepackage{graphicx}

\usepackage{tikz}
\usepackage{comment}
\usepackage{amsmath,amssymb} 
\usepackage{mathrsfs}

\usepackage{color}
\usepackage{subcaption}
\usepackage{multirow}
\usepackage{color}
\usepackage{booktabs}
\usepackage[multiple]{footmisc}

\usepackage[accsupp]{axessibility}  

\authorrunning{Huang et al.}

\begin{document}
\pagestyle{headings}
\mainmatter
\def\ECCVSubNumber{349}  

\title{FlowFormer: A Transformer Architecture for Optical Flow} 


\titlerunning{FlowFormer: A Transformer Architecture for Optical Flow}
%
\author{Zhaoyang Huang\inst{1,3}\thanks{Zhaoyang Huang and Xiaoyu Shi assert equal contributions.} \and
Xiaoyu Shi\inst{1,3}$^\star$ \and
Chao Zhang\inst{2} \and
Qiang Wang\inst{2} \and
Ka Chun Cheung\inst{3} \and
Hongwei Qin\inst{4} \and
Jifeng Dai\inst{4} \and
Hongsheng Li\inst{1}\thanks{Corresponding author: Hongsheng Li}
}
\institute{$^1$Multimedia Laboratory, The Chinese University of Hong Kong \and
$^2$Samsung Telecommunication Research \and
$^3$NVIDIA AI Technology Center \ \ \ \ 
$^4$SenseTime Research \\
\email{\{drinkingcoder@link, xiaoyushi@link,  hsli@ee\}.cuhk.edu.hk}
}
\maketitle

\begin{abstract}
We introduce optical Flow transFormer, dubbed as FlowFormer, a transformer-based neural network architecture for learning optical flow.
FlowFormer tokenizes the 4D cost volume built from an image pair, encodes the cost tokens into a cost memory with alternate-group transformer~(AGT) layers in a novel latent space, and decodes the cost memory via a recurrent transformer decoder with dynamic positional cost queries.
On the Sintel benchmark, FlowFormer achieves 1.159 and 2.088 average end-point-error~(AEPE) on the clean and final pass, a 16.5\% and 15.5\% error reduction from the best published result~(1.388 and 2.47). 
Besides, FlowFormer also achieves strong generalization performance.
Without being trained on Sintel, FlowFormer achieves 1.01 AEPE on the clean and pass of Sintel training set, outperforming the best published result~(1.29) by 21.7\%.


\end{abstract}

\section{Introduction}
Optical flow targets at estimating per-pixel correspondences between a source image and a target image, in the form of a 2D displacement field.
In many downstream video tasks, such as action recognition~\cite{sun2018optical,piergiovanni2019representation,zhao2020improved}, video inpainting~\cite{kim2019deep,xu2019deep,gao2020flow}, video super-resolution~\cite{lai2017deep,chan2021basicvsr,sajjadi2018frame}, and frame interpolation~\cite{xu2019quadratic,liu2020video,huang2020rife}, optical flow serves as a fundamental component providing dense correspondences as valuable clues for prediction.

A general assumption adopted in optical flow estimation is that the appearance of corresponding locations in the two images induced from optical flows remains unchanged. Traditionally, optical flow is modeled as an optimization problem that maximizes visual similarities between cross-image corresponding locations with regularization terms. With the rapid development of deep learning and emerging training data, this field has been significantly advanced by deep convolutional neural network-based methods. The recent methods compute costs (i.e. visual similarities) between feature pairs, upon which flows are regressed. Most successful architecture designs in optical flow are achieved via better designs of cost encoding and decoding. PWC-Net \cite{sun2018pwc} and RAFT \cite{teed2020raft} are two recent representative deep learning-based methods. PWC-Net \cite{sun2018pwc} builds hierarchical local cost volumes with warped features and progressively estimates flows from such local costs. RAFT \cite{teed2020raft} forms an $H\times W\times H\times W$ 4D cost volume that measures similarities between all pairs of pixels of the $H \times W$ image pair and iteratively retrieves local costs within local windows for regressing flow residuals.

Recently, transformers have attracted much attention for their ability of modeling long-range relations, which can benefit optical flow estimation. Perceiver IO \cite{jaegle2021perceiver} is the pioneering work that learns optical flow regression with a transformer-based architecture. However, it directly operates on pixels of image pairs and ignores the well-established domain knowledge of encoding visual similarities to costs for flow estimation. It thus requires a large number of parameters and $\sim80\times$ training examples to capture the desired input-output mapping. We therefore raise a question: can we enjoy both advantages of transformers and the cost volume from the previous milestones? Such a question calls for designing novel transformer architectures for optical flow estimation that can effectively aggregate information from the cost volume. In this paper, we introduce the novel optical Flow TransFormer~(FlowFormer) to address this challenging problem.

FlowFormer adopts an encoder-decoder architecture for cost volume encoding and decoding. After building a 4D cost volume, FlowFormer consists of two main components: 1) a cost volume encoder that embeds the 4D cost volume into a latent cost space and fully encodes the cost information in such a space, and 2) a recurrent cost decoder that estimates ﬂows from the encoded latent cost features. Compared with previous works, the main characteristic of our FlowFormer is to adapt the transformer architectures to effectively process cost volumes, which are compact yet rich representations widely explored in optical flow estimation communities, for estimating accurate optical flows.

A naive strategy to transform the 4D cost volume with transformers is directly tokenizing the 4D cost volume and applying transformers. However, such a strategy needs to use thousands of tokens, which is computationally unbearable. To tackle this challenge, we propose two key designs in our cost encoder.
We propose a two-step tokenization: 1) converting each of the 2D cost maps, which records visual similarities between one source pixel and all target pixels, from the 4D cost volume into patches as commonly done in transformer networks, and 2) further projecting cost-map patches of each cost map into $K$ latent cost tokens. In this way, 
the $H\times W \times H \times W$ 4D cost volume can be transformed into $H\times W\times K$ tokens.
Secondly, instead of performing self-attention among all tokens, we alternatively conduct attention over tokens within the same cost map and tokens across different cost maps. In other words, an interweaving stack of aggregations of latent cost tokens belonging to the same source pixel and those across different source pixels. 
Combining these two designs, FlowFormer encodes the cost volume into compact and globally aware latent cost tokens, dubbed as the {\it cost memory}.

Classical transformer architectures, such as DETR~\cite{carion2020end}, decodes information from the encoded memory via stacked cross-attention layers. In contrast to them, inspired by RAFT, our cost decoder adopts only a recurrent attention layer that formulates the cost decoding as a recurrent query process with dynamic positional cost queries: based on current estimated flows, we query the cost memory for regressing the flow residuals.
In each iteration, we compute the corresponding positions in the target image for all source pixels according to current flows and then dynamically update positional cost queries with such positions.
Then, we fetch cost features from the cost memory via cross-attention and use a shared gated recurrent unit~(GRU) head for residual flow regression. Moreover, RAFT only utilizes a shallow CNN as the image feature encoder.
We find that our FlowFormer can be benefited from using an ImageNet-pretrained transformer backbone.

Our contributions can be summarized as fourfold. 1) We propose a novel transformer-based neural network architecture, FlowFormer, for optical flow estimation, which achieves state-of-the-art flow estimation performance. 2) We design a novel cost volume encoder, effectively aggregating cost information into compact latent cost tokens.
3) We propose a recurrent cost decoder that recurrently decodes cost features with dynamic positional cost queries to iteratively refine the estimated optical flows.
4) To the best of our knowledge, we validate for the first time that an ImageNet-pretrained transformer can benefit the estimation of optical flow.

\section{Related Work}
\textbf{Optical Flow.} Traditionally, optical flow was modeled as an optimization problem that maximizes visual similarity between image pairs with regularizations \cite{horn1981determining,black1993framework,bruhn2005lucas,sun2014quantitative}. Major improvements in this era came from better designs of similarity and regularization terms.
The rise of deep neural networks significantly advanced this field. FlowNet \cite{dosovitskiy2015flownet} was the first end-to-end convolutional network for optical flow estimation. Its successive work, FlowNet2.0 \cite{ilg2017flownet}, adopted a stacked architecture with warping operation, performing on par with state-of-the-art (SOTA) methods. 
Then a series of works, represented by SpyNet \cite{ranjan2017optical}, PWC-Net \cite{sun2018pwc,sun2019models}, LiteFlowNet \cite{hui2018liteflownet,hui2020lightweight} and VCN \cite{yang2019volumetric}, employed coarse-to-fine and iterative estimation methodology. These models inherently suffered from missing small fast-motion objects in coarse stage. To remedy this issue, Teed and Deng \cite{teed2020raft} proposed RAFT~\cite{teed2020raft}, which performs optical flow estimation in a coarse-and-fine (i.e. multi-scale search window in each iteration) and recurrent manner.
Based on RAFT architecture, many works \cite{jiang2021learning,xu2021high,jiang2021learning2,zhang2021separable,hofinger2020improving} were proposed to either reduce the computational costs or improve the flow accuracy.
Recently, optical flow was extended to more challenging settings, such as low-light~\cite{zheng2020optical}, foggy~\cite{yan2020optical}, and lighting variations~\cite{huang2021life}.

Among these explorations, visual similarity is computed by the correlation of high dimensional features encoded by a convolutional neural network, and the cost volume that contains visual similarity of pixels pairs acts as a core component supporting optical flow estimation.
However, their cost information utilization lacks effectiveness.
We propose FlowFormer that aggregates the cost volume in a latent space with transformers~\cite{vaswani2017attention}.
Perceiver IO~\cite{jaegle2021perceiver} pioneered the use of transformers~\cite{vaswani2017attention,dosovitskiy2020image,carion2020end} that is able to establish long-range relationship in optical flow and achieved state-of-the-art performance. 
It ignored the cost volume, showing the strong expressive capacity of transformer architecture at the cost of $\sim80\times$ training examples.
In contrast, we propose to keep cost volume as a compact similarity representation and push search space to the extreme by globally aggregating similarity information via a transformer architecture.
Such global encoding operation is especially beneficial in the hard cases of large displacement and occlusion.

\noindent\textbf{Transformers for Computer Vision.} Transformers achieved great success in Natural Language Processing \cite{vaswani2017attention,dai2019transformer,devlin2018bert}, which inspired the development of self-attention for image classification~\cite{liu2021swin,dosovitskiy2020image,chu2021twins}.
Since then, transformer-based architectures has been introduced into many other vision tasks, such as detection \cite{carion2020end}, point cloud processing \cite{guo2021pct,zhao2021point}, image restoration \cite{chen2021pre,liang2021swinir}, video inpainting\cite{zeng2020learning,liu2021fuseformer}, visual grounding~\cite{yang2022improving}, etc, and achieves state-of-the-art in most tasks. The appealing performance is generally attributed to the long-range modeling capacity, which is also a desired property in optical flow estimation.
One of the challenges that vision transformers are faced with is the large number of visual tokens because the computational cost quadratically increases along with the token number.
Twins~\cite{chu2021twins} proposed a spatially separable self-attention (SS Self-Attention) layer that propagates information over tokens arranged in a 2D plane.
We also adopt the SS Self-Attention in the cost volume encoder to propagate information inter-cost-maps.
Perceiver IO~\cite{jaegle2021perceiver} proposed a general transformer backbone, which although requires a large amount of parameters, achieves state-of-the-art optical flow performance.
Visual correspondence tasks~\cite{sun2021loftr,huang2021vs,cho2021cats,jiang2021cotr,xu2022rnnpose} is a main stream in computer vision.
Recently, transformers also lead a trend in such tasks~\cite{sarlin2020superglue,sun2021loftr,cho2021cats,jiang2021cotr}, which is more related to ours.

\section{Method}
\begin{figure}[t!]
    \centering
    \resizebox{0.95\linewidth}{!}{
        \includegraphics[width=\linewidth, trim={0mm 40mm 75mm 5mm}, clip]{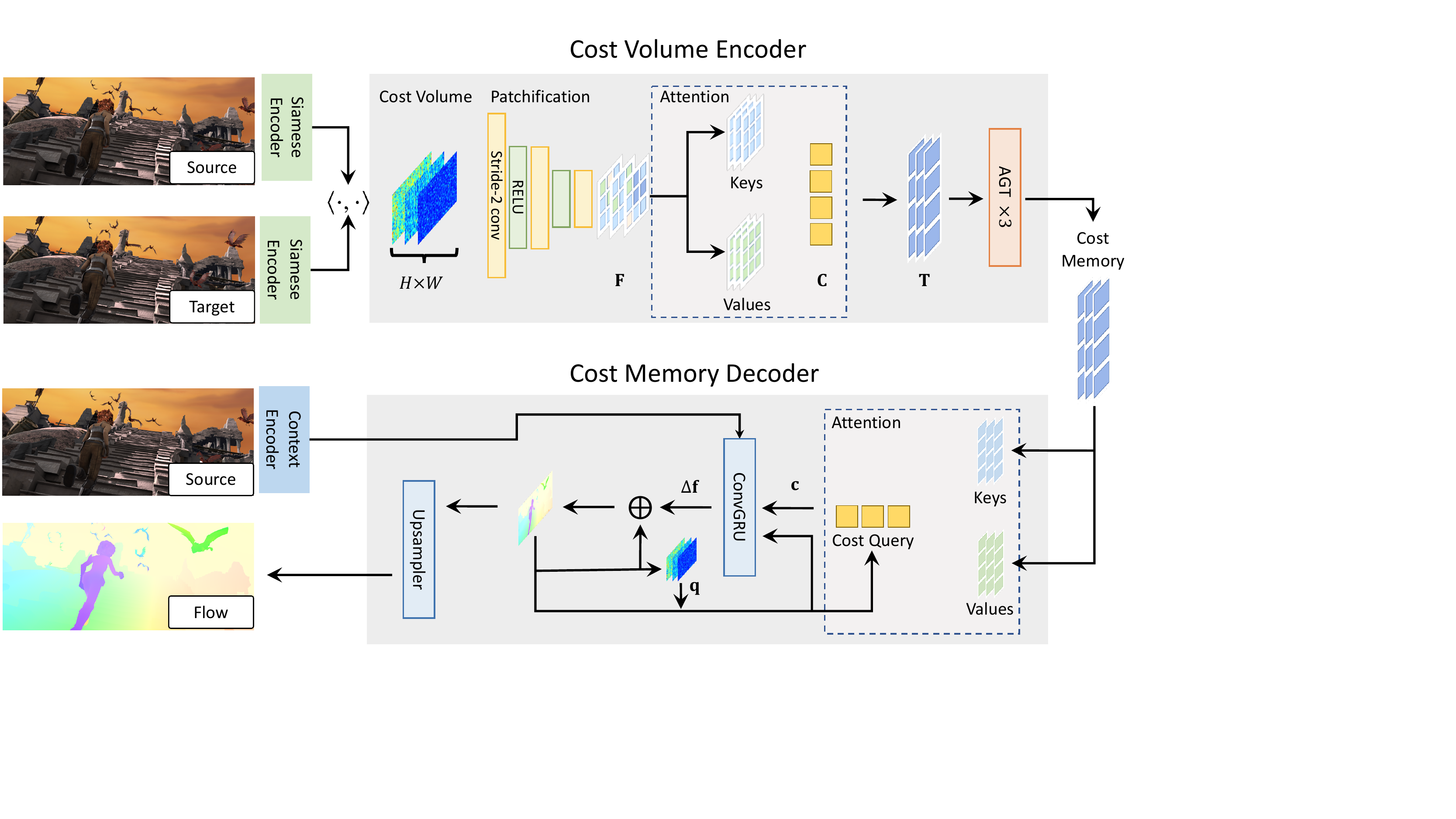}
    }
    \caption{Architecture of FlowFormer. FlowFormer estimates optical flow in three steps: 1) building a 4D cost volume from image features.
    2) A cost volume encoder that encodes the cost volume into the cost memory. 3) A recurrent transformer decoder that decodes the cost memory with the source image context features into flows.
    }
    \label{Fig: arc}
\end{figure}

The task of optical flow estimation requires to output a per-pixel displacement field $\mathbf{f}: \mathbb{R}^{2} \rightarrow \mathbb{R}^2$ that maps every 2D location $\mathbf{x} \in \mathbb{R}^{2}$ of a source image $\mathbf{I}_s$ to its corresponding 2D location $\mathbf{p} = \mathbf{x} + \mathbf{f}(\mathbf{x})$ of a target image $\mathbf{I}_t$. 
To take advantage of the recent vision transformer architectures as well as the 4D cost volumes widely utilized by previous CNN-based optical flow estimation methods,
we propose FlowFormer, a transformer-based architecture that encodes and decodes the 4D cost volume to achieve accurate optical flow estimation.
In Fig.~\ref{Fig: arc}, we show the overview architecture of FlowFormer, which processes the 4D cost volumes from siamese features with two main components: 1) a cost volume encoder that encodes the 4D cost volume into a latent space to form cost memory, and 2) a cost memory decoder for predicting a per-pixel displacement field based on the encoded cost memory and contextual features.

\subsection{Building the 4D Cost Volume}

A backbone vision network is used to extract an $H\times W\times D_f$ feature map from an input $H_I\times W_I\times 3$ RGB image, where typically we set $(H, W)=(H_I/8, W_I/8)$.
After extracting the feature maps of the source image and the target image, we construct an $H\times W\times H\times W$ 4D cost volume by computing the dot-product similarities between all pixel pairs between the source and target feature maps.

\subsection{Cost Volume Encoder}

To estimate optical flows, the corresponding positions in the target image of source pixels need to be identified based on source-target visual similarities encoded in the 4D cost volume.
The built 4D cost volume can be viewed as a series of 2D cost maps of size $H \times W$, each of which measures visual similarities between a single source pixel and all target pixels.
We denote source pixel $\bf x$'s cost map as ${\bf M_x} \in \mathbb{R}^{H \times W}$.
Finding corresponding positions in such cost maps is generally challenging, as there might exist repeated patterns and non-discriminative regions in the two images. 
The task becomes even more challenging when only considering costs from a local window of the map, as previous CNN-based optical flow estimation methods do.
Even for estimating a single source pixel's accurate displacement, it is beneficial to take its contextual source pixels' cost maps into consideration.

To tackle this challenging problem, we propose a transformer-based cost volume encoder that encodes the whole cost volume into a {\it cost memory}.
Our cost volume encoder consists of three steps: 1) cost map patchification, 2) cost patch token embedding, and 3) cost memory encoding.
We elaborate the details of the three steps as follows.

\noindent \textbf{Cost map patchification.} 
Following existing vision transformers, we patchify the cost map $\mathbf{M_x} \in \mathbb{R}^{H \times W}$ of each source pixel $\bf x$ with strided convolutions to obtain a sequence of cost patch embeddings.
Specifically, given an $H \times W$ cost map, we first pad zeros at its right and bottom sides to make its width and height multiples of 8. 
The padded cost map is then transformed by a stack of three stride-2 convolutions followed by ReLU into a feature map $\mathbf{F}_{\bf x} \in \mathbb{R}^{\lceil{H/8}\rceil \times \lceil{W/8}\rceil \times D_p}$.
Each feature in the feature map stands for an $8\times 8$ patch in the input cost map.
The three convolutions have output channels of $D_p/4$, $D_p/2$, $D_p$, respectively.

\noindent \textbf{Patch Feature Tokenization via Latent Summarization.}
Although the patchification results in a sequence of cost patch feature vectors for each source pixel, the number of such patch features is still large and hinders the efficiency of information propagation among different source pixels. Actually, a cost map is highly redundant because only a few high costs are most informative. To obtain more compact cost features, we further summarize the patch features $\mathbf F_{\bf x}$ of each source pixel $\mathbf x$ via $K$ latent codewords $\mathbf{C} \in \mathbb{R}^{K\times D}$. Specifically, the latent codewords query each source pixel's cost-patch features to further summarize each cost map into $K$ latent vectors of $D$ dimensions via the dot-product attention mechanism. The latent codewords ${\bf C} \in \mathbb{R}^{K \times D}$
are randomly initialized, updated via back-propagation, and shared across all source pixels. The latent representations ${\bf T_x}$ for summarizing ${\bf F_x}$ are obtained as
\begin{equation}
\begin{aligned}
    & {\bf K_x} = \text{Conv}_{1\times 1}\left(\text{Concat}({\bf F_x}, {\rm PE})\right), \\
    & {\bf V_x} = \text{Conv}_{1\times 1}\left(\text{Concat}({\bf F_x}, {\rm PE})\right), \\
    & {\bf T_x} = \text{Attention}({\bf C}, {\bf K_x}, {\bf V_x}).
\end{aligned}
\label{Eq: latent projection}
\end{equation}
Before projecting the cost-patch features $\bf F_x$ to obtain keys $\bf K_x$ and values $\bf V_x$, the patch features are concatenated with a sequence of positional embeddings $\text{PE} \in \mathbb{R}^{\lceil{H/8}\rceil\times \lceil{W/8}\rceil\times D_p}$. 
Given a 2D position $\mathbf p$, we encode it into a positional embedding of length $D_p$ following COTR \cite{jiang2021cotr}.
Finally, the cost map of the source pixel $\bf x$ can be summarized into $K$ latent representations ${\bf T_x} \in \mathbb{R}^{K\times D}$ by conducting multi-head dot-product attention with the queries, keys, and values. Generally, $K \times D \ll H \times W$ and the latent summarizations $\bf T_x$ therefore provides more compact representations than each $H \times W$ cost map for each source pixel $\bf x$.
For all source pixels in the image, there are a total of $(H\times W)$ 2D cost maps. Their summarized representations can consequently be converted into a latent 4D cost volume ${\bf T} \in \mathbb{R}^{H \times W \times K \times D}$.

\begin{figure}[t!]
    \centering
    \resizebox{1.0\linewidth}{!}{
        \includegraphics[width=\linewidth, trim={10mm 125mm 30mm 0}, clip]{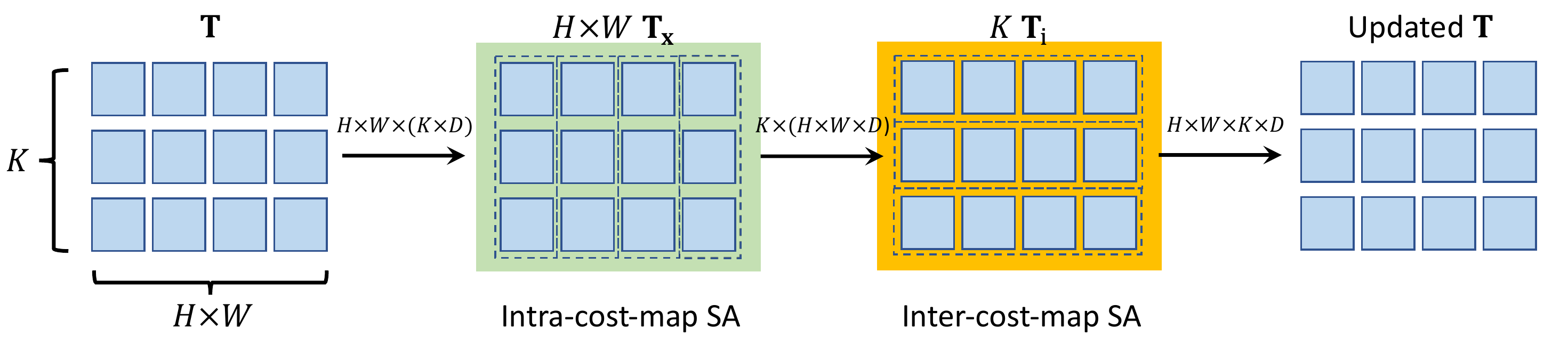}
    }
    \caption{Alternate-Group Transformer Layer. The alternate-group transformer layer~(AGT) alternatively groups tokens in $\mathbf T$ into $H\times W$ groups that contains $K$ tokens ($\mathbf T_\mathbf x$) and $K$ groups that contains $H\times W$ tokens ($\mathbf T_i$), and encode tokens inside groups via self-attention and ss self-attention~\cite{chu2021twins} respectively.
    }
    \label{Fig: agt}
\end{figure}

\noindent \textbf{Attention in the latent cost space.} The aforementioned two stages transform the original 4D cost volume into a latent and compact 4D cost volume $\bf T$. 
However, it is still too expensive to directly apply self-attention over all the vectors in the 4D volume because the computational cost quadratically increases with the number of tokens. 
As shown in Fig.~\ref{Fig: agt}, we propose an alternate-group transformer layer (AGT) that groups the tokens in two mutually orthogonal manners and apply attentions in the two groups alternatively, which reduces the cost of attention while still being able to propagate information among all tokens.

The first grouping is conducted for each source pixel, i.e., each ${\bf T_x} \in \mathbb{R}^{K \times D}$ forms a group and the self-attention is conducted within each group.
\begin{equation}
\begin{aligned}
    {\bf T_x} = {\rm FFN}({\rm Self\text{-}Attention}({\bf T_x}(1), \dots, {\bf T_x}(K)) ~~~\text{ for all } {\bf x} \text{ in } {\bf I}_s,
\end{aligned}
\label{Eq: intra-cost}
\end{equation}
where ${\bf T_x}(i)$ denotes the $i$-th latent representation for encoding the source pixel $\bf x$'s cost map. After the self-attention is conducted between all $K$ latent tokens for each source pixel $\bf x$, updated $\bf T_x$ are further transformed by a feed-forward network (FFN) and then re-organized back to form the updated 4D cost volume $\bf T$. Both the self-attention and FFN sub-layers adopt the common designs of residual connection and layer normalization of transformers.
This self-attention operation propagates the information within each cost map and we name it as intra-cost-map self-attention.

The second way groups all the latent cost tokens ${\bf T} \in \mathbb{R}^{H \times W \times K \times D}$
into $K$ groups according to the $K$ different latent representations. 
Each group would therefore have $(H \times W)$ tokens of dimension $D$ for information propagation in the spatial domain via
the spatially separable self-attention (SS-SelfAttention) proposed in Twins~\cite{chu2021twins},
\begin{equation}
\begin{aligned}
    {\bf T}_i = {\rm FFN}({\rm SS}\text{-}{\rm SelfAttention}({\bf T}_i)) ~~~\text{ for } i = 1,2,\dots, K,
\end{aligned}
\label{Eq: inter-cost}
\end{equation}
where we slightly abuse the notation and denote ${\bf T}_i \in \mathbb{R}^{(H \times W) \times D}$ as the $i$-th group. The updated ${\bf T}_i$'s are then re-organized back to obtain the updated 4D latent cost volume $\bf T$. Moreover, visually similar source pixels should have coherent flows, which has been validated by previous methods~\cite{jiang2021learning,cho2021cats}.
Thus, we integrate appearance affinities between different source pixels into SS-SelfAttention via concatenating the source image's context features $\mathbf{t}$ with the cost tokens when generating queries and keys.
We call this layer inter-cost-map self-attention layer as it propagates information of cost volume across different source pixels.
Note that these two operations are different from CATs~\cite{cho2021cats}, which augmented correlations ‘intra’ a level of cost
map and ‘inter’ multi-level correlation layers.

The above self-attention operations' parameters are shared across different groups and they are sequentially operated to form the proposed alternate-group attention layer.
By stacking the alternate-group transformer layer multiple times, the latent cost tokens can effectively exchange information across source pixels and across latent representations to better encode the 4D cost volume.
In this way, our cost volume encoder transforms the $H\times W\times H\times W$ 4D cost volume to $H\times W\times K$ latent tokens of length $D$.
We call the final $H\times W\times K$ tokens as the \textit{cost memory}, which is to be decoded for optical flow estimation.

\subsection{Cost Memory Decoder for Flow Estimation}

Given the cost memory encoded by the cost volume encoder, we propose a cost memory decoder to predict optical flows. 
Since the original resolution of the input image is $H_I \times W_I$,
we estimate optical flow at the $H\times W$ resolution and then upsample the predicted flows to the original resolution with a learnable convex upsampler \cite{teed2020raft}.
However, in contrast to previous vision transformers that seek abstract semantic features, optical flow estimation requires recovering dense correspondences from the cost memory.
Inspired by RAFT \cite{teed2020raft}, we propose to use cost queries to retrieve cost features from the cost memory and
iteratively refine flow predictions with a recurrent attention decoder layer.

\noindent \textbf{Cost memory aggregation.}
For predicting the flows of the $H\times W$ source pixels, we generate a sequence of $(H \times W)$ cost queries, each of which is responsible for estimating the flow of a single source pixel via co-attention on the cost memory. To generate the cost query $\bf Q_x$ for a source pixel $\bf x$, we first compute its corresponding location in the target image given its current estimated flow ${\bf f}({\bf x})$ as ${\bf p} = {\bf x} + {\bf f}({\bf x})$. 
We then retrieve a local $9\times 9$ cost-map patch $\mathbf{q_x}=\mathrm{Crop}_{9\times 9}(\mathbf{M_x}, \mathbf{p})$ by cropping costs inside the $9\times 9$ local window centered at $\bf p$ on the cost map $\mathbf{M_x}$.
The cost query $\bf Q_x$ is then formulated based on the features $\text{FFN}(\mathbf{q_x})$ that encoded from the local costs $\mathbf{q_x}$ and $\bf p$'s positional embedding $\mathrm{PE}({\bf p})$, which can aggregate information from source pixel $\bf x$'s cost memory $\bf T_x$ via cross-attention,
\begin{equation}
\begin{aligned}
    & {\bf Q_x} = \text{FFN}\left(\text{FFN}(\mathbf{q_x})  + \mathrm{PE}(\mathbf p) \right), \\
    & {\bf K_x} = \text{FFN}\left({\bf T_x}\right),~~{\bf V_x} = \text{FFN}\left({\bf T_x}\right), \\
    & {\bf c_x} = \mathrm{Attention}({\bf Q_x}, {\bf K_x}, {\bf V_x}).
\end{aligned}
\label{Eq: decoder attention}
\end{equation}
The cross-attention summarizes information from the cost memory for each source pixel to predict its flow. 
As $\bf Q_x$ is dynamically updated in terms of the fed position at each iteration, we call it as dynamic positional cost query.
We note that keys and values can be generated at the beginning and re-used in subsequent iterations, which saves computation as a benefit of our recurrent decoder.

\noindent \textbf{Recurrent flow prediction.} 
Our cost decoder iteratively regresses flow residuals $\Delta \mathbf f(\mathbf x)$ to refine the flow of each source pixel $\mathbf x$ as $\mathbf f(\mathbf x) \leftarrow \mathbf f(\mathbf x) + \Delta \mathbf f(\mathbf x)$.
We adopt a $\text{ConvGRU}$ module and follow the similar design to that in GMA-RAFT~\cite{jiang2021learning} for flow refinement. 
However, the key difference of our recurrent module is the use of cost queries to adaptively aggregate information from the cost memory for more accurate flow estimation.
Specifically, at each iteration, the ConvGRU unit takes as input the concatenation of retrieved cost features and cost-map patch $\mathrm{Concat}({\bf c_x, q_x})$, the source-image context feature $\bf t_x$ from the context network, and the current estimated flow $\mathbf f$, and outputs the predicted flow residuals as follows,
\begin{equation}
\begin{aligned}
    & \Delta \mathbf f(\mathbf x) = \text{ConvGRU}(\mathrm{Concat}({\bf c_x}, {\bf q_x}), {\bf t_x}, \mathbf f(\mathbf x)).
\end{aligned}
\label{Eq: delta flow}
\end{equation}
The flows generated at each iteration are unsampled to the size of the source image via a convex upsampler following \cite{teed2020raft} and supervised by ground-truth flows at all recurrent iterations with increasing weights.

\section{Experiment}
We evaluate our FlowFormer on the Sintel~\cite{butler2012naturalistic} and the KITTI-2015~\cite{geiger2013vision} benchmarks.
Following previous works, we train FlowFormer on FlyingChairs~\cite{dosovitskiy2015flownet} and FlyingThings~\cite{mayer2016large}, and then respectively finetune it for Sintel and KITTI benchmark.
Flowformer achieves state-of-the-art performance on both benchmarks.

\noindent \textbf{Experimental setup}.
We use the average end-point-error~(AEPE) and F1-All(\%) metric for evaluation.
The AEPE computes mean flow error over all valid pixels.
The F1-all, which refers to the percentage of pixels whose flow error is larger than 3 pixels or over 5\% of length of ground truth flows.
The Sintel dataset is rendered from the same model but in two passes, i.e. clean pass and final pass.
The clean pass is rendered with smooth shading and specular reflections.
The final pass uses full rendering settings including motion blur, camera depth-of-field blur, and atmospheric effects.

\noindent \textbf{Implementation details}.
The image feature encoder of our final FlowFormer is chosen as the first two stages of ImageNet-pretrained Twins-SVT~\cite{chu2021twins}, which encodes an image into $D_f=256$-channel feature map of 1/8 image size. The cost volume encoder patchifies each cost map to a $D_p=64$-channel feature map and further summarizes the feature map to $N=8$ cost tokens of $K=128$ dimensions.
Then, the cost volume encoder encodes the cost tokens with 3 AGT layers.
Following previous optical flow training procedure~\cite{jiang2021learning}, we pre-train FlowFormer on FlyingChairs~\cite{dosovitskiy2015flownet} for 120k iterations with a batch size of 8, and on FlyingThings~\cite{mayer2016large} for 120k iterations with a batch size of 6~(denoted as `C+T').
After pre-training, we finetune FlowFormer on the data combined from FlyingThings, Sintel, KITTI-2015, and HD1K~\cite{kondermann2016hci}~(denoted as `C+T+S+K+H') for 120k iterations with a batch size of 6.
To achieve the best performance on the KITTI benchmark, we also further finetune FlowFormer on the KITTI-2015 for 50k iterations with a batch size of 6.
We use the one-cycle learning rate scheduler. The highest learning rate is set as $2.5\times 10^{-4}$ on FlyingChairs and $1.25\times 10^{-4}$ on the other training sets.
As positional encodings used in transformers are sensitive to image size, we crop the image pairs for flow estimation and tile them to obtain complete flows following Perceiver IO~\cite{jaegle2021perceiver}. 
We use the tile technique for evaluating optical flow on KITTI because the size of images in KITTI is quite different from training image size.
We use fixed Gaussian weights for tile, which will be detailed in the supplementary materials.

\subsection{Quantitative Experiment}

\begin{table}[t]
\centering
\resizebox{1.0\linewidth}{!}{
\begin{tabular}{clccccccc}
\multicolumn{1}{c}{\multirow{2}{*}{Training Data}} & \multicolumn{1}{c}{\multirow{2}{*}{Method}} & \multicolumn{2}{c}{Sintel (train)}                    & \multicolumn{2}{c}{KITTI-15 (train)}                    & \multicolumn{2}{c}{Sintel (test)}                     & \multicolumn{1}{c}{KITTI-15 (test)} \\
\cmidrule(r{1.0ex}){3-4}\cmidrule(r{1.0ex}){5-6}\cmidrule(r{1.0ex}){7-8} \cmidrule(r{1.0ex}){9-9} 
\multicolumn{1}{c}{}                               & \multicolumn{1}{c}{}                        & \multicolumn{1}{c}{Clean} & \multicolumn{1}{c}{Final} & \multicolumn{1}{c}{F1-epe} & \multicolumn{1}{c}{F1-all} & \multicolumn{1}{c}{Clean} & \multicolumn{1}{c}{Final} & \multicolumn{1}{c}{F1-all} \\ 
\hline
\multirow{3}{*}{A}                         &    Perceiver IO~\cite{jaegle2021perceiver}   & 1.81 & 2.42 & 4.98  & - & - & - & - \\ 
&    PWC-Net~\cite{sun2018pwc}   & 2.17 & 2.91 & 5.76 & - & - & - & -  \\ 
& RAFT~\cite{teed2020raft}  & 1.95 & 2.57 &  4.23 & - & - & - & - \\
\hline
\multicolumn{1}{c}{\multirow{9}{*}{C+T}}           & HD3~\cite{yin2019hierarchical} &  3.84      & 8.77      & 13.17  & 24.0  & - & - & - \\
 & LiteFlowNet~\cite{hui2018liteflownet} & 2.48 & 4.04 & 10.39 & 28.5 & - & - & -  \\
 & PWC-Net~\cite{sun2018pwc} & 2.55 & 3.93 & 10.35 & 33.7 & - & - & - \\
 & LiteFlowNet2~\cite{hui2020lightweight} & 2.24 & 3.78 & 8.97 & 25.9 & - & - & - \\
 & S-Flow~\cite{zhang2021separable}  & 1.30 & 2.59 & 4.60 & 15.9 & & & \\
 & RAFT~\cite{teed2020raft}   & 1.43 & 2.71 & 5.04 & 17.4 & - & - & - \\
 & FM-RAFT~\cite{jiang2021learning2}   & 1.29 & 2.95 & 6.80 & 19.3 & - & - & -  \\
 & GMA~\cite{jiang2021learning}   & 1.30 & 2.74 & 4.69 & 17.1 & - & - & - \\
 & Ours   & $\mathbf{1.01}$ & $\mathbf{2.40}$ & $\mathbf{4.09}^{\dag}$ & $\mathbf{14.72}^{\dag}$ & - & - & - \\
\hline
\multirow{11}{*}{C+T+S+K+H}                         &    LiteFlowNet2~\cite{hui2020lightweight}    &    (1.30)                       &       (1.62)                    &   (1.47)                          &      (4.8)                      &       3.48                     &         4.69                   &          7.74   \\
 &  PWC-Net+~\cite{sun2019models} & (1.71) & (2.34) &  (1.50) &  (5.3) & 3.45 &  4.60 & 7.72\\
 & VCN~\cite{yang2019volumetric} & (1.66) & (2.24) & (1.16) & (4.1) & 2.81 & 4.40 & 6.30 \\
 &  MaskFlowNet~\cite{zhao2020maskflownet} & - & - & - & - & 2.52 & 4.17 & 6.10 \\
  &  S-Flow~\cite{zhang2021separable} & (0.69) & (1.10) & (0.69) & (1.60) & 1.50 & 2.67 & $\mathbf{4.64}$ \\
 &  RAFT~\cite{teed2020raft} & (0.76) & (1.22) & (0.63) & (1.5) & 1.94 & 3.18 & 5.10
\\
 &  FM-RAFT~\cite{jiang2021learning2} & (0.79) & (1.70) & (0.75) & (2.1) & 1.72 & 3.60 &  6.17 \\
 &  GMA~\cite{jiang2021learning} & - & - & - & - & 1.40 & 2.88 & 5.15 \\
&  Ours & (0.48) & (0.74) & (0.53) & (1.11) & $\mathbf{1.16}$ & $\mathbf{2.09}$ & $4.68^\dag$  \\
\cmidrule(r{1.0ex}){2-9}
&  RAFT*~\cite{teed2020raft} & (0.77) & (1.27) & - & - & 1.61 & 2.86 & -
\\
 &  GMA*~\cite{jiang2021learning} & (0.62) & (1.06) & (0.57) & (1.2) & 1.39 & 2.47 & - \\
\hline
\end{tabular}
}
\caption{\label{Tab: comparison} Experiments on Sintel~\cite{butler2012naturalistic} and KITTI~\cite{geiger2013vision} datasets. `A' denotes the autoflow dataset. `C + T' denotes training only on the FlyingChairs and FlyingThings datasets. `+ S + K + H' denotes finetuning on the combination of Sintel, KITTI, and HD1K training sets.  * denotes that the methods use the warm-start strategy~\cite{teed2020raft}, which relies on previous image frames in a video. $^\dag$ is estimated via the tile technique elaborated in the supplementary.
Our FlowFormer achieves best generalization performance~(C+T) and ranks 1st on the Sintel benchmark~(C+T+S+K+H).
}
\end{table}

We evaluate FlowFormer on the well-known Sintel and KITTI benchmarks as shown in Tab.~\ref{Tab: comparison}.
GMA~\cite{jiang2021learning}, an improved version of RAFT~\cite{teed2020raft}, is the most competitive flow estimation method at present.
After being trained on FlyingChairs and FlyingThings, we evaluate the generalization performance of FlowFormer on the training set of Sintel and KITTI-2015.
By further finetuning FlowFormer on the combination of HD1K, Sintel and KITTI training sets,
we compare the dataset-specific accuracy of optical flow models.
Autoflow~\cite{sun2021autoflow} is a dataset that provides training data covering various challenging visual disturbance, but its training code is not released yet.

\noindent \textbf{Generalization performance.}
We train FlowFormer on the FlyingChairs and FlyingThings~(C+T), and evaluate it on the training set of Sintel and KITTI-2015. This settings evaluates the generalization performance of optical flow models.
FlowFormer ranks 1st among all compared methods on both benchmarks.
FlowFormer achieves 1.01 and 2.40 on the clean and final pass of Sintel.
On the KITTI-2015 training set, FlowFormer achieves 4.09 F1-epe and 14.72 F1-all.
Compared to GMA, FlowFormer reduces 22.3\% and 12.4\% errors on Sintel clean and final, and 13.9\% errors on KITTI-2015 F1-all, which shows its extraordinary generalization performance.
RAFT trained on the autoflow dataset~(A) significantly outperforms RAFT trained on the C+T on final pass because autoflow provides training image pairs that are more challenging.
We believe training FlowFormer with autoflow can achieve better accuracy but it is not released yet.

\noindent \textbf{Sintel benchmark.}
We finetune the pretrained FlowFormer on the combination of training data of FlyingThings, HD1K, Sintel and KITTI-2015, and then evaluate it on the Sintel test set.
FlowFormer achieves 1.16 and 2.09 on the Sintel clean and final, 16.5\% and 15.5\% lower error compared to GMA$^*$, which ranks both 1st on the Sintel benchmark.
It is noteworthy that RAFT$^*$ and GMA$^*$ use the warm-start strategy that requires image sequences while FlowFormer does not.
Compared with GMA, which also does not use the warm-start, FlowFormer obtains 17.2\% and 27.5\% error reduction.

\noindent \textbf{KITTI-2015 benchmark.}
We further finetune the FlowFormer on the KITTI-2015 training set after the Sintel finetuning stage and evaluate it on the KITTI test set.
FlowFormer achieves 4.68, ranking 2nd on the KITTI-2015 benchmark.
S-Flow~\cite{zhang2021separable} obtains slightly smaller error than FlowFormer on KITTI~($-$0.85\%), which, however, is significantly worse on Sintel~(31.6\% and 22.5\% larger error on clean and final pass).
S-Flow finds corresponding points by computing the coordinate expectation weighted by refined cost maps.
Images in the KITTI dataset are captured in urban traffic scenes, which contains objects that are mostly rigid.
Flows on rigid objects are rather simple, which is easier for cost-based coordinate expectation, but the assumption can be easily violated in non-rigid scenarios such as Sintel.

\begin{figure}[t!]
    \centering
    \resizebox{1.0\linewidth}{!}{
    \begin{subfigure}[t]{0.33\linewidth}
        \includegraphics[width=\linewidth]{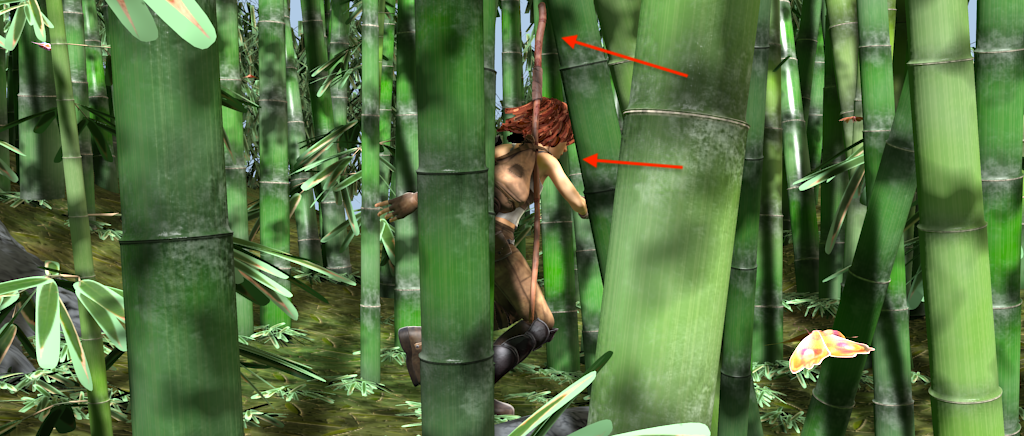}
    \end{subfigure}
    \begin{subfigure}[t]{0.33\linewidth}
        \includegraphics[width=\linewidth]{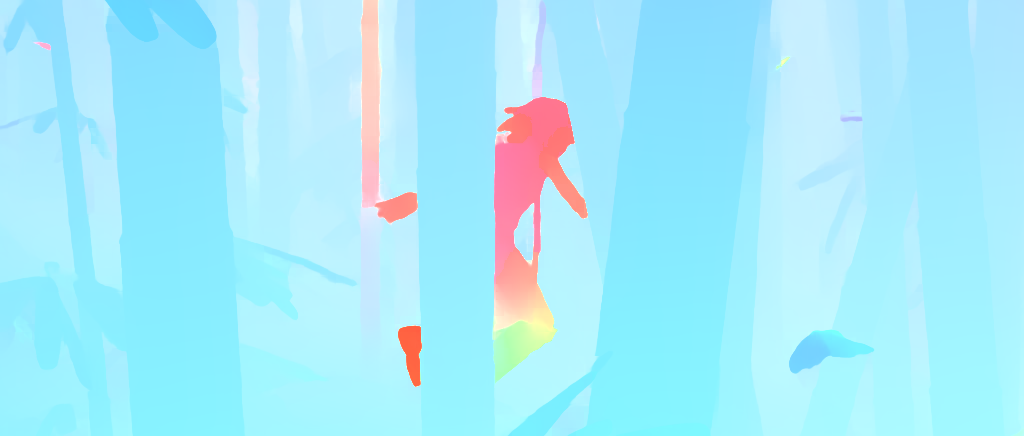}
    \end{subfigure}
    \begin{subfigure}[t]{0.33\linewidth}
        \includegraphics[width=\linewidth]{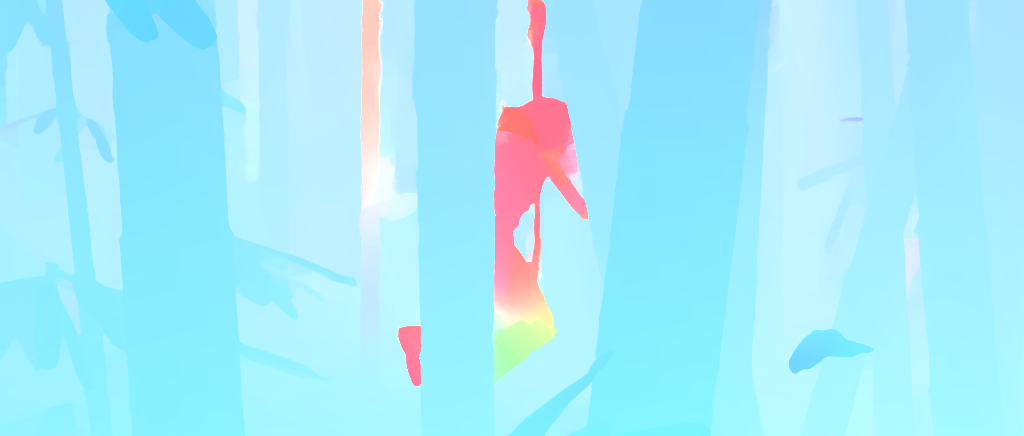}
    \end{subfigure}
    }
    \resizebox{1.0\linewidth}{!}{
    \begin{subfigure}[t]{0.33\linewidth}
        \includegraphics[width=\linewidth]{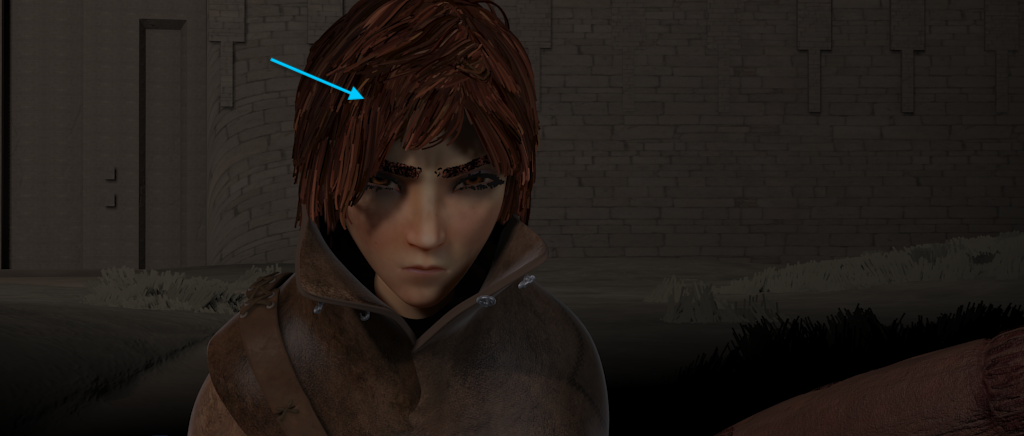}
    \end{subfigure}
    \begin{subfigure}[t]{0.33\linewidth}
        \includegraphics[width=\linewidth]{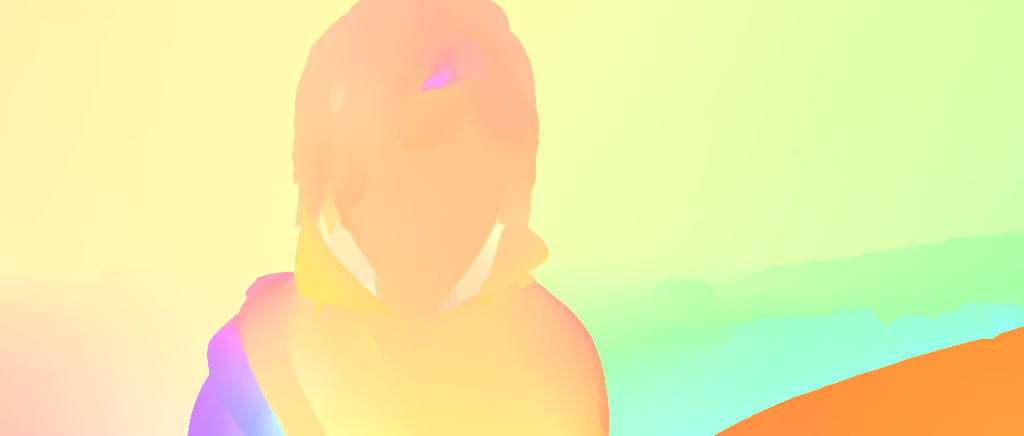}
    \end{subfigure}
    \begin{subfigure}[t]{0.33\linewidth}
        \includegraphics[width=\linewidth]{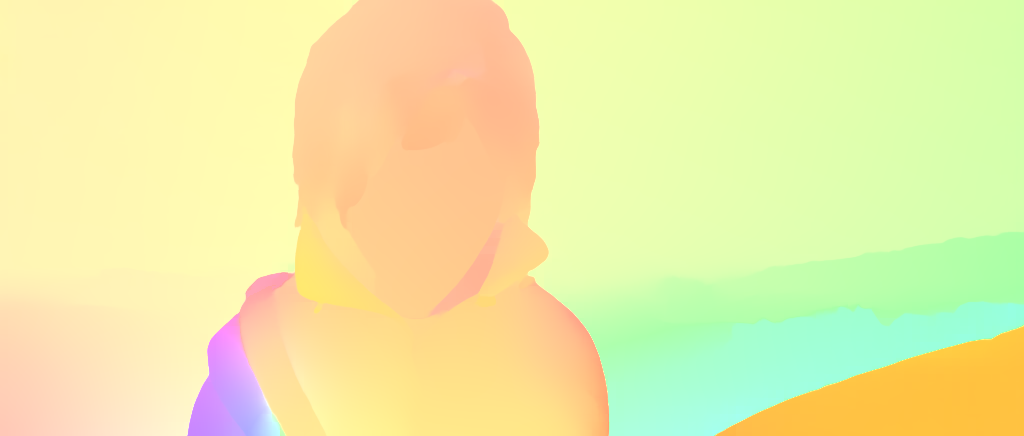}
    \end{subfigure}
    }
    \resizebox{1.0\linewidth}{!}{
    \begin{subfigure}[t]{0.33\linewidth}
        \includegraphics[width=\linewidth]{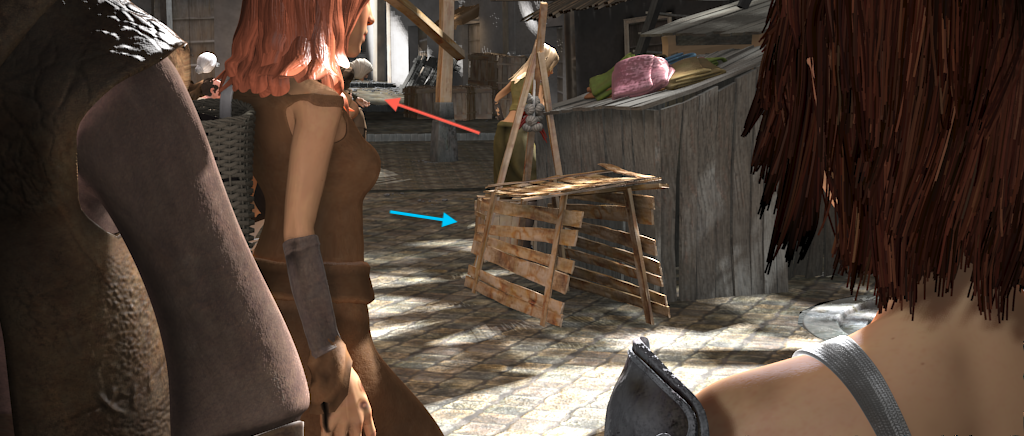}
        \caption{\small Input}
    \end{subfigure}
    \begin{subfigure}[t]{0.33\linewidth}
        \includegraphics[width=\linewidth]{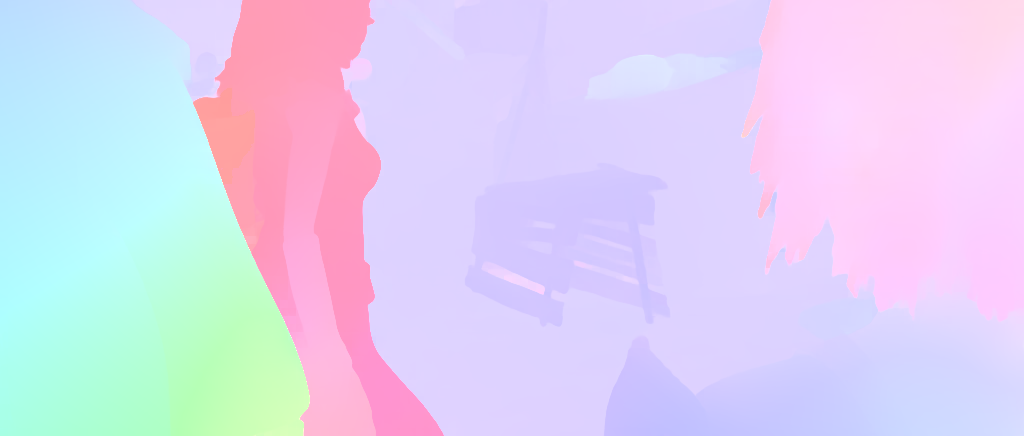}
        \caption{\small FlowFormer (Ours)}
    \end{subfigure}
    \begin{subfigure}[t]{0.33\linewidth}
        \includegraphics[width=\linewidth]{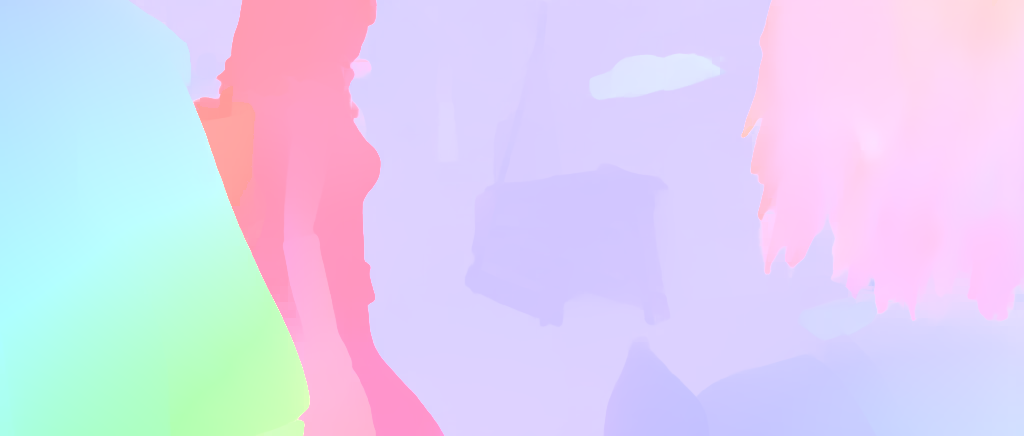}
        \caption{\small GMA}
    \end{subfigure}
    }
    \caption{Qualitative comparison on the Sintel test set. FlowFormer greatly reduces the flow leakage around object boundaries~(pointed by red arrows) and clearer details~(pointed by blue arrows).
    }
    \label{Fig: flow comparison}
\end{figure}

\subsection{Qualitative Experiment}

We visualize flows that estimated by our FlowFormer and GMA of three examples in Fig.~\ref{Fig: flow comparison} to qualitatively show how FlowFormer outperforms GMA.
As transformers can encode the cost information at a large perceptive field, FlowFormer can distinguish overlapping objects via contextual information and thus reduce the leakage of flows over boundaries.
Compared with GMA, the flows that are estimatd by FlowFormer on boundaries of the bamboo and the human body are more precise and clear.
Besides, FlowFormer can also recover motion details that are ignored by GMA, such as the hair and the holes on the box.

\subsection{Ablation Study}

\begin{table}[t]
\centering
\resizebox{0.92\linewidth}{!}{
\begin{tabular}{clccccc}
\hline
\multicolumn{1}{c}{\multirow{2}{*}{Experiment}} & \multicolumn{1}{c}{\multirow{2}{*}{Method}} & \multicolumn{2}{c}{Sintel (train)}                    & \multicolumn{2}{c}{KITTI-15 (train)}                 & \multirow{2}{*}{Params.}  \\
\cmidrule(r{1.0ex}){3-4}\cmidrule(r{1.0ex}){5-6}
\multicolumn{1}{c}{}                               & \multicolumn{1}{c}{}                        & \multicolumn{1}{c}{Clean} & \multicolumn{1}{c}{Final} & \multicolumn{1}{c}{F1-epe} & \multicolumn{1}{c}{F1-all} &    \\ 
\hline
\hline
baseline & RAFT  & 1.53 & 2.99 & 5.73 & 18.29 & 5.3M    \\ 
\hline
\multirow{3}{*}{MCR$\rightarrow$LCT+CMD}                       
& $K=4$, $D=32$ & 1.66 & 2.93 & 5.60 & 19.67 &  5.5M \\
& $K=8$, $D=32$   & 1.58 & 2.90 & 5.50 & 18.71 &  5.5M  \\ 
 & $K=8$, $D=128$ & 1.44 & 2.80 & 5.22 & 17.64 &  5.6M  \\ 
\hline
\multirow{3}{*}{CNN $\rightarrow$ Twins}
 & CNN & 1.44 & 2.80 & 5.22 & 17.64 &  5.6M  \\ & Twins from Scratch & 1.44 & 2.86 & 5.38 & 17.58 &  14.0M \\
& Pretrained Twins    & 1.29 & 2.72 & 4.82 & 16.16 &   14.0M  \\ 
\hline
\multirow{5}{*}{Cost Encoding}                        
& None    & 1.29 & 2.72 & 4.82 & 16.16 &   14.0M  \\ 
& +Intra. & 1.29 & 2.89 & 4.74 & 15.71 &  14.1M \\
& AGT$\times$1 (+Intra.+Inter.) & 1.20 & 2.85 & 4.57 & 15.46 & 15.2M  \\ 
& AGT$\times$2 & 1.16 & 2.66 & 4.70 & 16.01 & 16.4M  \\ 
& AGT$\times$3 & 1.10 & 2.57 & 4.45 & 15.15 & 17.6M  \\ 
\hline
\end{tabular}
}
\caption{\label{Tab: ablation} Ablation study. We gradually change one component of the RAFT at a time to obtain our FlowFormer model. MCR$\rightarrow$LCT+CMD: replacing RAFT's decoder with OUR latent cost tokens + cost memory decoder. CNN$\rightarrow$Twins: replacing RAFT's CNN encoder with Twins-SVT transformer. Cost Encoding: adding intra-cost-map and inter-cost-map to form an Alternate-Group Transformer layer in the encoder. 3 AGT layers are used in our final model.
}
\end{table}

\begin{table}[h]
\centering
\resizebox{0.92\linewidth}{!}{
\begin{tabular}{lccccccc}
& \multicolumn{1}{c}{\multirow{2}{*}{Method}} & \multicolumn{2}{c}{Sintel (train)}                    & \multicolumn{2}{c}{KITTI-15 (train)} & \multirow{2}{*}{parameters} \\
\cmidrule(r{1.0ex}){3-4}\cmidrule(r{1.0ex}){5-6}
\multicolumn{1}{c}{}                               & \multicolumn{1}{c}{}                        & \multicolumn{1}{c}{Clean} & \multicolumn{1}{c}{Final} & \multicolumn{1}{c}{F1-epe} & \multicolumn{1}{c}{F1-all} & \\ 
\hline
 & GMA~\cite{jiang2021learning}   & 1.30~(+30\%) & 2.74~(+12\%) & 4.69~(+15\%) & 17.1~(+16\%) & 5.9M \\
 & Ours~(small)  & 1.20~(+20\%) & 2.64~(+8\%) & 4.57~(+12\%) & 16.62~(+13\%) & 6.2M \\
 \cline{2-7}
 & GMA-L~\cite{jiang2021learning}   & 1.33~(+33\%) & 2.56~(+4\%) & 4.40~(+8\%) & 15.93~(+8\%) & 17.0M \\
 & GMA-Twins~\cite{jiang2021learning}   & 1.15~(+15\%) & 2.73~(+11\%) & 4.98~(+22\%) & 16.82~(+14\%) & 14.2M \\
 & Ours   & $\mathbf{1.00}$ & $\mathbf{2.45}$ & $\mathbf{4.09}$ & $\mathbf{14.72}$ & 18.2M
\\ \hline
\end{tabular}
}
\caption{\label{Tab: param comparison} FlowFormer v.s. GMA. 
Ours~(small) is a small version of FlowFormer and uses the CNN image feature encoder of GMA.
GMA-L is a large version of GMA. 
GMA-Twins replace its CNN image feature encoder with pre-trained Twins.
(+x\%) indicates that this model obtains x\% larger error than ours.
}
\end{table}

We conduct a series of ablation experiments in Tab.~\ref{Tab: ablation}.
We start from RAFT as the baseline, which directly regresses residual flows with the multi-level cost retrieval~(MCR) decoder,
and gradually replace its components with our proposed components.
We first replace RAFT's MCR decoder with the latent cost tokenization (LCT) part of our encoder and the cost memory decoder (CMD) (denoted as `MCR$\rightarrow$LCT+CMD'). Note that our cost memory decoder cannot be used alone on top of the 4D cost volume of RAFT because of the too large number of tokens. It must be combined with our latent cost tokens ($\bf T_x$ from Eq. (1)). Encoding $K=8$ latent tokens of $D=128$ dimensions for each source pixel achieves the best performance.
Based on LCT+CMD with $K=8$ and $D=128$, we replace RAFT's CNN image feature encoder with Twins-SVT~(denoted as `CNN$\rightarrow$Twins').
We then further add attention layers of the proposed cost volume encoder to encode and update latent cost tokens.
The proposed Alternate-Group Transformer (AGT) layer consists of two types of attention, i.e., intra-cost-map attention and inter-cost-map attention.
We first add a single intra-cost-map attention layer (denoted as `+Intra.'), and then add the inter-cost-map attention (denoted as `AGT$\times1$ (+Intra.+Inter.)', which is equivalent to adding a single AGT layer.
We then test on increasing the number of AGT layers to 2 and 3.
Following RAFT, all models are trained on FlyingChairs~\cite{dosovitskiy2015flownet} with 100k iterations and FlyingThings~\cite{mayer2016large} with 60k iterations, and then evaluated on the training set of Sintel~\cite{butler2012naturalistic} and KITTI-2015~\cite{geiger2013vision}.

\noindent \textbf{MCR $\rightarrow$ LCT+MCD.}
The number of latent tokens $K$ and token dimension $D$ determine how much cost volume information the cost tokens can encode.
From $K=4, D=32$ to $K=8, D=128$, the AEPE decreases because the cost tokens summarizes more cost map information and benefits the residual flow regression.
The latent cost tokens are capable of summarizing whole-image information and our MCD can absorb interested information from them through co-attention, while the MCR decoder of RAFT only retrieves multi-level costs inside flow-guided local windows.
Therefore, even without our AGT layers in our encoder,  LCT+MCD still shows better performance than MCR decoder of RAFT.

\noindent \textbf{CNN vs. Transformer Image Encoder.}
In the CNN$\rightarrow$Twins experiment, 
the AEPE of Twins trained from scratch is marginally worse than CNN, but the ImageNet-pretraining is beneficial,
because Twins is a transformer architecture with larger receptive field and model capacity, which requires more training examples for sufficient training. 

\noindent \textbf{Cost Encoding.} 
In the cost volume encoder, we encode and update the latent cost tokens with an intra-cost-map attention operation and an inter-cost-map attention operation. The two operations form an Alternate-Group Transformer (AGT) layer.
Then we gradually increase the number of AGT layers to 3.
From no attention layer to AGT$\times$3, the errors gradually decrease, which demonstrates that encoding latent cost tokens with our AGT layers benefits flow estimation.

\noindent \textbf{FlowFormer vs. GMA.} We train all the models with the settings of GMA. The full version of FlowFormer has 18.2M parameters, which is larger than GMA. One of the causes is that FlowFormer uses the first two stages of ImageNet-pretrained Twins-SVT as the image feature encoder while GMA uses a CNN.
We present an experiment to compare FlowFormer and GMA with aligned settings in Tab.~\ref{Tab: param comparison}.
We first provide a small version of FlowFormer using GMA's CNN image encoder and also set $K=4$, $D=32$, and AGT$\times$1.
Although the smaller version of FlowFormer (denoted as 'Ours (small)') has a significant performance drop compared to the full version of FlowFormer, it still outperforms GMA in terms of all metrics.
We also design two enhanced GMA models and compare them with the full version of FlowFormer to show that the performance improvements are not simply derived from adding more parameters.
The first one is denoted as `GMA-L', a large version of GMA and the second one is denoted as `GMA-Twins' which also adopts the pretrained Twins as the image encoder. 
In this experiment, we train all models on FlyingChairs with 120k iterations and FlyingThings with 120k iterations.
Similar to reducing RAFT to RAFT~(small)~\cite{teed2020raft}, GMA-L enlarges GMA by doubling feature channels, which has 17M parameters, comparable to FlowFormer.
However, its performance degrades in Sintel clean, a 33\% larger error than FlowFormer.
GMA-Twins replaces the CNN image encoder with the shallow Image-Net pre-trained Twins-SVT as FlowFormer does.
The largest improvement of GMA-Twins upon GMA is on the Sintel clean, but it still has a 15\% larger error than FlowFormer. GMA-Twins does not lead to significant error reduction on other metrics and is even worse on the KITTI-15.
In conclusion, the performance improvement of FlowFormer is not derived from more parameters but the novel design of the architecture.

\section{Conclusion}
We have proposed FlowFormer, a Transformer-based architecture for optical flow estimation.
FlowFormer summarizes the $H\times W\times H\times W$ 4D cost volume built from a pair of images as $H\times W\times K$ tokens of length $D$, and then efficiently and effectively encodes the cost tokens via the alternate-group transformer~(AGT).
Thanks to such design, the generated cost memory is able to grasp essential information over the cost volume and obtain compact cost features.
Finally, the cost memory decoder absorbs cost information from the cost memory with dynamic positional cost queries, which gets rid of the limitation of local windows, for residual flow regression.
To our best knowledge, FlowFormer is the first method that deeply integrates transformers with cost volumes for optical flow estimation.
Thanks to the compact cost tokens and long-range relation modeling ability of transformers, FlowFormer achieves state-of-the-art accuracy and shows strong cross-dataset generalization.

\noindent \textbf{Acknowledgements.}
Hongsheng Li is also a Principal Investigator of Centre for Perceptual and Interactive Intelligence Limited (CPII). This work is supported in part by CPII, in part by the General Research Fund through the Research Grants Council of Hong Kong under Grants (Nos. 14204021, 14207319), in part by CUHK Strategic Fund. Thanks Qiaole Dong and Deqing Sun for the code check.



\clearpage
%
%
\bibliographystyle{splncs04}
\bibliography{egbib}
\end{document}


\pagestyle{headings}
\mainmatter
\def\ECCVSubNumber{349}  

\title{FlowFormer: A Transformer Architecture for Optical Flow -- Supplementary Materials} 


\titlerunning{FlowFormer -- Supplementary Materials}
%
\author{Zhaoyang Huang\inst{1,3}\thanks{Zhaoyang Huang and Xiaoyu Shi assert equal contributions.} \and
Xiaoyu Shi\inst{1,3}$^\star$ \and
Chao Zhang\inst{2} \and
Qiang Wang\inst{2} \and
Ka Chun Cheung\inst{3} \and
Hongwei Qin\inst{4} \and
Jifeng Dai\inst{4} \and
Hongsheng Li\inst{1}\thanks{Corresponding author: Hongsheng Li}
}
\institute{$^1$Multimedia Laboratory, The Chinese University of Hong Kong \and
$^2$Samsung Telecommunication Research \and
$^3$NVIDIA AI Technology Center \ \ \ \ 
$^4$SenseTime Research \\
\email{\{drinkingcoder@link, xiaoyushi@link,  hsli@ee\}.cuhk.edu.hk}
}
\maketitle



%

\section{More Ablation Studies}

\begin{table}[h]
\centering
\begin{tabular}{llllllll}
\hline
  & \multirow{2}{*}{Intra.} & \multirow{2}{*}{Inter.} & \multicolumn{2}{l}{Sinte (train)} & \multicolumn{2}{l}{Kitti (train)} & \multirow{2}{*}{Params.} \\
  &  &  & Clean & Final & F1-epe & F1-all & \\ 
\hline
 & Trans & Trans & 1.20 & 2.85 & 4.57 & 15.46 & 15.2M \\
 \hline
 & MLP & Trans & 1.20 & 2.67 & 5.01 & 16.81 & 15.2M \\
& Trans & Conv & 1.23 & 2.72 & 4.73 & 15.87 & 15.1M \\
& MLP & Conv & 1.22 & 2.71 & 4.88 & 17.23 & 15.1M \\
\hline
\end{tabular}
\caption{\label{Tab: ablation} Ablation study on the alternative-group transformer~(AGT) layer. For intra-cost-map aggregation layer~(Intra.), we replace transformer (Trans) with MLP-Mixer~\cite{tolstikhin2021mlpmixer} block (MLP). For inter-cost-map aggregation layer~(Inter.), we replace transformer with ConvNeXt~\cite{Liu2022ACF} block (Conv).
}
\end{table}

As shown in Table \ref{Tab: ablation}, we conduct additional ablation experiments on the alternative-group transformer (AGT) layer. For intra-cost-map aggregation layer, since the number and dimension of latent cost tokens are fixed, we test on replacing our design with MLP-Mixer~\cite{tolstikhin2021mlpmixer}~(2nd row), which is a state-of-the-art MLP-based architecture. We also substitute  ConvNeXt~\cite{Liu2022ACF} for transformer in inter-cost-map aggregation~(3rd row).
Furthermore, we replace both transformers with MLP and ConvNext~(4th row).
Replacing transformer layers leads to slightly better performance on Sintel final pass, while brings a clear drop on KITTI.
Therefore, we adopt the proposed full transformer architecture as our final model.

\section{Tile with Gaussian Weights}

Since positional encodings used in transformers are sensitive to image size and the size of an image pair for test $(H_{test}\times W_{test})$ might be different from those of the training images, $(H_{train}\times W_{train})$,
we crop the test image pair according to the training size and estimate flows for patch pairs separately, and then tile the flows to obtain a complete flow map following a similar strategy proposed in Perceiver IO \cite{jaegle2021perceiver}.
Specifically, we crop the image pair into four evenly-spaced tiles, i.e., $H_{train}\times W_{train}$ image tiles starting at $(0, 0)$, $(0, W_{test}-W_{train})$, $(H_{test}-H_{train}, 0)$, and $(H_{test}-H_{train}, W_{test}-W_{train})$, respectively. For each pixel that is covered by several tiles, we compute its output flow $\bf f$ by blending the predicted flows ${\bf f}_i$ with weighted averaging:
\begin{equation}
\begin{aligned}
    {\bf f} = \frac{\sum_i  w_i {\bf f}_i}{\sum_i w_i},
\end{aligned}
\label{Eq: weighted average}
\end{equation}
where $w_i$ is the weight of the $i$-th tile for the pixel.
We compute the $H_{train}\times W_{train}$ weight map according to pixels' normalized distances $d_{u, v}$ to the tile center:
\begin{equation}
\begin{aligned}
& d_{u,v}=||(u/H_{train}-0.5, v/W_{train}-0.5)||_2, \\
& w_{u,v}=g(d_{u,v};\mu=0, \sigma=0.05),
\end{aligned}
\label{Eq: gaussian}
\end{equation}
where $(u, v)$ denote a pixel's 2D coordinate.
We use a Gaussian-like $g$ as the weighting function to obtain smoothly blended results. We use this weight map for all the tiles.

\section{Training Image Size Details}
We train FlowFormer with image size of $368\times 498$ on FlyingChairs and $432\times960$ on the following training stages, i.e., FlyingThings, Sintel, and KITTI.
As the height of images in KITTI only ranges from 374 to 375, we train another FlowFormer model, dubbed as \textit{FlowFormer\#}, and evaluate it on the KITTI-15 training set to obtain better performance.  Following GMA~\cite{jiang2021learning}, \textit{FlowFormer\#} is trained with $368\times 498$ image size on FlyingChairs and $400\times 720$ image size on FlyingThings, which achieves 4.09 F1-epe and 14.72 F1-all on the KITTI training set as presented in the Table 1 in the original paper.

\clearpage
%
%



\clearpage
%
%
\bibliographystyle{splncs04}
\bibliography{egbib}